\documentclass{article}
\usepackage{spconf,amsmath,graphicx}
\usepackage{algorithm,algpseudocode}
\usepackage{mathtools}
\usepackage{cancel}
\usepackage{amssymb}
\usepackage{booktabs}
\usepackage{caption}
\usepackage{subcaption}
\usepackage{multirow}
\usepackage{setspace}
\usepackage{icomma}

\usepackage{pifont}
\usepackage{tikz}
\usetikzlibrary{positioning}
\usetikzlibrary{calc}
\usepackage{pgfplots, pgfplotstable}    
\pgfplotsset{compat=1.17} 
\usepackage{url}
\DeclareMathSymbol{\shortminus}{\mathbin}{AMSa}{"39}
\usepackage{xcolor}
\usepackage{enumitem}
\usepackage{hyperref}
\hypersetup{
    colorlinks=true,
    linkcolor=purple,
    filecolor=magenta,      
    urlcolor=purple,
}
\usepackage[capitalize]{cleveref}

\makeatletter
\def\blfootnote{\xdef\@thefnmark{}\@footnotetext}
\makeatother

\usepackage[
backend=biber,
style=ieee,
citestyle=numeric-comp,
maxbibnames=3,
maxcitenames=3,
doi=false,isbn=false,url=false,eprint=false
]{biblatex}

\defbibheading{bibliography}[\refname]{}

\DeclareSourcemap{
	\maps[datatype=bibtex, overwrite=true]{
		\map{
			\step[fieldsource=booktitle,
			match=\regexp{.*Interspeech.*},
			replace={Proc. Interspeech}]
			\step[fieldsource=journal,
			match=\regexp{.*INTERSPEECH.*},
			replace={Proc. Interspeech}]
			\step[fieldsource=booktitle,
			match=\regexp{.*ICASSP.*},
			replace={Proc. ICASSP}]
			\step[fieldsource=booktitle,
			match=\regexp{.*icassp_inpress.*},
			replace={Proc. ICASSP (in press)}]
			\step[fieldsource=booktitle,
			match=\regexp{.*Acoustics,.*Speech.*and.*Signal.*Processing.*},
			replace={Proc. ICASSP}]
			\step[fieldsource=booktitle,
			match=\regexp{.*International.*Conference.*on.*Learning.*Representations.*},
			replace={Proc. ICLR}]
			\step[fieldsource=booktitle,
			match=\regexp{.*International.*Conference.*on.*Computational.*Linguistics.*},
			replace={Proc. COLING}]
			\step[fieldsource=booktitle,
			match=\regexp{.*SIGdial.*Meeting.*on.*Discourse.*and.*Dialogue.*},
			replace={Proc. SIGDIAL}]
			\step[fieldsource=booktitle,
			match=\regexp{.*International.*Conference.*on.*Machine.*Learning.*},
			replace={Proc. ICML}]
			\step[fieldsource=booktitle,
			match=\regexp{.*North.*American.*Chapter.*of.*the.*Association.*for.*Computational.*Linguistics:.*Human.*Language.*Technologies.*},
			replace={Proc. NAACL}]
			\step[fieldsource=booktitle,
			match=\regexp{.*Empirical.*Methods.*in.*Natural.*Language.*Processing.*},
			replace={Proc. EMNLP}]
			\step[fieldsource=booktitle,
			match=\regexp{.*Association.*for.*Computational.*Linguistics.*},
			replace={Proc. ACL}]
			\step[fieldsource=booktitle,
			match=\regexp{.*Automatic.*Speech.*Recognition.*and.*Understanding.*},
			replace={Proc. ASRU}]
			\step[fieldsource=booktitle,
			match=\regexp{.*Spoken.*Language.*Technology.*},
			replace={Proc. SLT}]
			\step[fieldsource=booktitle,
			match=\regexp{.*Speech.*Synthesis.*Workshop.*},
			replace={Proc. SSW}]
			\step[fieldsource=booktitle,
			match=\regexp{.*workshop.*on.*speech.*synthesis.*},
			replace={Proc. SSW}]
			\step[fieldsource=booktitle,
			match=\regexp{.*Advances.*in.*neural.*information.*processing.*},
			replace={Proc. NeurIPS}]
			\step[fieldsource=booktitle,
			match=\regexp{.*Advances.*in.*Neural.*Information.*Processing.*},
			replace={Proc. NeurIPS}]
			\step[fieldsource=booktitle,
			match=\regexp{.*Workshop.*on.* Applications.* of.* Signal.*Processing.*to.*Audio.*and.*Acoustics.*},
			replace={Proc. WASPAA}]
			\step[fieldsource=publisher,
			match=\regexp{.+},
			replace={{}}]
			\step[fieldsource=month,
			match=\regexp{.+},
			replace={{}}]
			\step[fieldsource=location,
			match=\regexp{.+},
			replace={{}}]
			\step[fieldsource=address,
			match=\regexp{.+},
			replace={{}}]
			\step[fieldsource=organization,
			match=\regexp{.+},
			replace={{}}]
		}
	}
}

\title{Exploring Speech Recognition, Translation, and Understanding with Discrete Speech Units: A Comparative Study}
\name{
\begin{tabular}{c}
Xuankai Chang$^1$, Brian Yan$^1$, Kwanghee Choi$^1$, Jeeweon Jung$^1$, Yichen Lu$^1$, Soumi Maiti$^1$,\\Roshan Sharma$^1$, Jiatong Shi$^1$, Jinchuan Tian$^1$, Shinji Watanabe$^{1\dagger}$, Yuya Fujita$^2$, Takashi Maekaku$^2$,\\Pengcheng Guo$^3$, Yao-Fei Cheng$^4$, Pavel Denisov$^5$, Kohei Saijo$^6$, Hsiu-Hsuan Wang$^7$\sthanks{Authors are ordered by the organizations. $^\dagger$Corresponding author.}
\end{tabular}}
\address{
$^1$Carnegie Mellon University, $^2$Yahoo Japan Corporation, $^3$Northwestern Polytechnical University, \\$^4$University of Washington,$^5$University of Stuttgart, $^6$Waseda University,$^7$National Taiwan University
}

\begin{document}
\ninept
\maketitle
%


\begin{abstract}
Speech signals, typically sampled at rates in the tens of thousands per second, contain redundancies, evoking inefficiencies in sequence modeling.
High-dimensional speech features such as spectrograms are often used as the input for the subsequent model. However, they can still be redundant.
Recent investigations proposed the use of discrete speech units derived from self-supervised learning representations, which significantly compresses the size of speech data.
Applying various methods, such as de-duplication and subword modeling, can further compress the speech sequence length. 
Hence, training time is significantly reduced while retaining notable performance.
In this study, we undertake a comprehensive and systematic exploration into the application of discrete units within end-to-end speech processing models.
Experiments on 12 automatic speech recognition, 3 speech translation, and 1 spoken language understanding corpora demonstrate that discrete units achieve reasonably good results in almost all the settings.
We intend to release our configurations and trained models to foster future research efforts.
\end{abstract}


\section{Introduction} \label{sec:intro}

Significant progress has been made in the field of automatic speech recognition (ASR) over the past few decades, largely attributed to the evolution of deep neural networks~\cite{hinton2012deep,qian2016very}.
Since the emergence of end-to-end (E2E) ASR models~\cite{graves2006connectionist,graves2012sequence,chorowski2015attention}, there have been many exciting outcomes.
Among these achievements, a slew of potent architectures~\cite{vaswani2017attention,gulati2020conformer,guo2021recent,kim2023branchformer} has boosted the performance of speech tasks, including ASR, speech translation (ST), and spoken language understanding (SLU).
Also, novel training paradigms have demonstrated improved performance and generalization, including self-supervised learning (SSL) models~\cite{schneider2019wav2vec,baevski2019vq,baevski2020wav2vec,hsu2021hubert,chen2022wavlm} and Whisper~\cite{radford2023robust}.
In the majority of prior endeavors, high-dimensional features are derived from raw waveforms as the input.
Conventionally, spectral speech features are extracted from a fixed-length temporal window, such as Mel Frequency Cepstral Coefficients (MFCC) or log Mel filter banks (FBANK).
Recently, data-driven methods become popular to learn feature extraction using neural networks~\cite{sainath2015learning,baevski2020wav2vec,hsu2021hubert}. However, in most cases, the data storage and transmission efficiency are similar among raw waveforms and speech features~\cite{chang2023exploration}.
It is not trivial to improve the efficiency of computation without performance degradation.

\begin{figure}[t]
    \centering
    \includegraphics[width=0.95\linewidth]{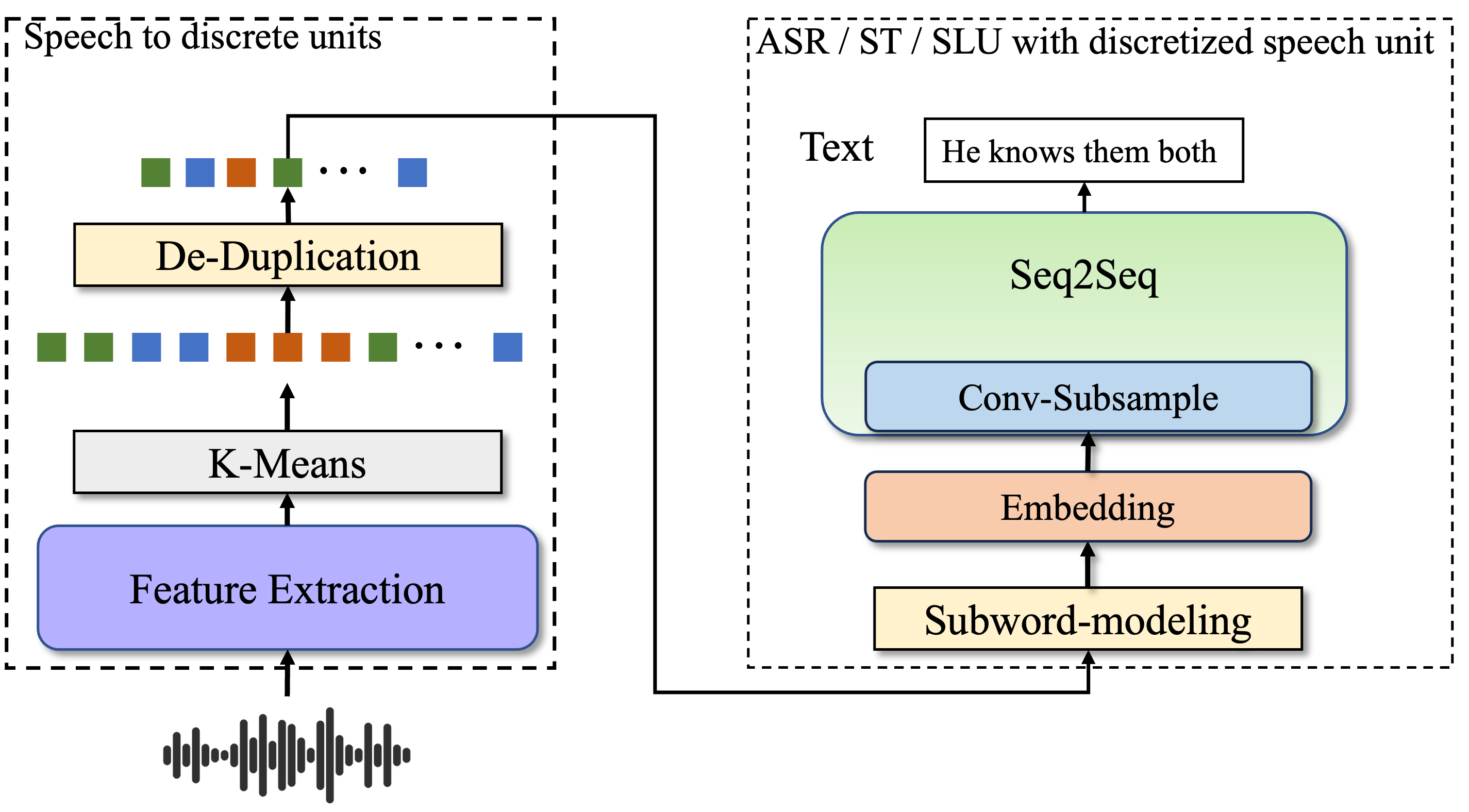}
    \vspace{-1em}
    \caption{Illustration of the E2E speech processing model with discrete speech units. The speech discretization process is shown on the left. On the right side is a Seq2Seq model that takes discrete units to output target text.}
    \label{fig:model_arch}
    \vspace{-1em}
\end{figure}

Recently, a few studies have proposed the use of discrete speech units to represent speech signals, where the information of speech signals in a short window is represented by a single token or a few levels from restricted vocabulary~\cite{baevski2019vq,baevski2020effectiveness,chang2023exploration}, as opposed to employing high-dimensional continuous speech features within the ASR task.
For instance, in~\cite{chang2023exploration}, researchers propose to utilize clustering indices derived from features of SSL models as input.
This approach condenses the information considerably, reducing the original 1024-sized float vector to a mere 12-bit binary number: over 3000 times less.
Such a compression process substantially reduces data storage and transmission size while maintaining predictive performance comparable to conventional high-dimensional features.
Moreover, input sequence lengths can be drastically reduced via de-duplication and subword modeling on discrete units, leading to more than 2X faster training and inference.
Notably, discrete unit representations can be regarded as a spoken language similar to text data in NLP tasks, which is more straightforward to unify the tasks or models.
Being an alternative to traditional speech representation, this paradigm has already been applied in other domains, such as speech translation~\cite{zhang2023dub,kim2023many} and audio generation~\cite{hayashi2020discretalk,shi2021discretization,borsos2023audiolm,wang2023neural,rubenstein2023audiopalm}.

To provide an extensive guide for discrete speech units, we conducted a comprehensive exploration into the effectiveness of discrete speech units across various speech processing tasks.
We summarize our experiments and findings as follows:
\begin{itemize}[leftmargin=*]
    \item A comparative analysis is conducted under fair conditions, evaluating the performance and training time reduction using discrete speech units as opposed to traditional speech features.
    \item A diverse range of benchmarks, including 12 ASR (\Cref{ssec:asr_results}), 3 ST (\Cref{ssec:st_results}), and 1 SLU (\Cref{ssec:slu_results}) corpora, are mostly evaluated for the first time. 
    \item To demonstrate wide applicability of the discrete units, we adopted noisy speech, spontaneous speech, telephony speech, and several multi-lingual speech corpora, which would be the first work to explore these aspects (\Cref{sssec:asr_datasets}).
    \item We show the versatility of discrete units in various E2E frameworks, including connectionist temporal classification (CTC)~\cite{graves2006connectionist}, attention-based encoder-decoder (AED)~\cite{chorowski2015attention}, and RNN-Transducer~\cite{graves2012sequence} (\Cref{tab:asr_ctc_transducer_cer_wer}). 
    \item We share various tips based on our investigations to get better performance, including SSL feature choice and discretization. Selecting SSL features based on canonical correlation analysis (CCA)~\cite{pasad2023comparative} improves performance significantly compared to prior work~\cite{chang2023exploration} (\Cref{ssec:exp_setup}).
    \item We also explore other possible choices of discrete units, including clustering SSL \cite{chen2022wavlm} or supervised representations, or vector quantization of neural codec models \cite{defossez2022high} (\Cref{sssec:asr_different_units}). 
    \item We will release fully reproducible recipes and trained models on ESPnet \cite{espnet}, which can significantly benefit the community.
\end{itemize}
\section{Speech Processing with Discrete Tokens} \label{sec:discrete_token_sp}
This section elaborates on the details of our speech processing models, which take discrete units as input.
\Cref{fig:model_arch} summarizes the whole pipeline.
Leveraging the sequence-to-sequence (Seq2Seq) paradigm as our backbone framework enables broad applicability across a spectrum of tasks involving the transformation of speech signals into diverse target outputs.

\subsection{Speech Discretization} \label{ssec:speech_disc}
The discretization process is the pivotal step, which transforms the speech signals into discrete representations.
Previous studies have employed techniques such as vector quantization (VQ) and clustering for discretization.
The former typically necessitates the learning of a specialized vector quantization module during the training phase.
Prominent examples include VQ-VAE~\cite{van2017neural}, Wav2Vec models~\cite{schneider2019wav2vec,baevski2019vq,baevski2020wav2vec}, and recent popular neural codec models~\cite{zeghidour2021soundstream,defossez2022high} with multi-level residual VQ component.
The alternative applies the clustering algorithms to the features extracted from the signals, where the cluster indices are directly used as the discrete representations.
K-Means clustering is often used in HuBERT-like models~\cite{hsu2021hubert,chen2022wavlm}.

We are in favor of the clustering-based methods due to their inherent versatility, thereby catering to diverse tasks.
The benefit of this approach comes in threefolds:
\begin{enumerate}[leftmargin=*]
    \item It enables a wide choice of feature extraction methods, including spectral features or intermediate representations from SSL or supervised learning-based models.
    \item Distinct layers retain different information \cite{chen2022wavlm}, and we can choose an optimal feature for different purposes.
    \item The vocabulary size can be easily tuned for the balance of information distinctions and efficiency without modifying the pre-trained models.
\end{enumerate}

\subsection{Data Manipulations} \label{ssec:token_mani}
The discrete units derived from the preceding stages are temporally aligned with the original speech features.
However, the discrete units still contain trivial redundancies of speech, such as repeated or commonly co-existing units.
Note that after the speech signal is converted into discrete units, it can be regarded as a type of spoken language similar to tokenized text in conventional NLP tasks.
Hence, similar processing and modeling techniques can be applied in a trivial way.
To remove the redundancies introduced by repeated or commonly co-existing units, two methods have been employed to significantly reduce the input sequence length~\cite{chang2023exploration}:

\begin{enumerate}[leftmargin=*]
\item \textbf{De-duplication}: This approach involves condensing consecutive subsequences featuring identical tokens into a single token to reduce redundancy. 
\item \textbf{Subword Modeling}: This technique combines frequent patterns of discrete unit subsequences and reassigns them to metatokens, to enhance the input token representation.
\end{enumerate}

Besides these two methods, simple data augmentation, \textbf{time masking}, is also adopted in~\cite{chang2023exploration}, serving as a regularization technique during training.

\subsection{Seq2Seq Modeling with Convolutional Subsampling} \label{ssec:seq2seq_model}
We use Seq2Seq models to map the discrete speech units to other target outputs with E2E training.
We take the advantage of well-established Seq2Seq models, including AED~\cite{chorowski2015attention}, CTC~\cite{graves2006connectionist}, and RNN-Transducer~\cite{graves2012sequence}.
The detailed definitions of these models can be found in their references.

\noindent\textbf{Convolutional Subsampling.}
The input sequence length of discrete tokens is generally longer than the target sequence length.
As we employ the widely used recent self-attention model \cite{kim2023branchformer} in our experiments, the sequence length poses a problem.
The computation cost is quadratically dependent on the input sequence length.
Hence, to reduce the computation overhead, we introduce a convolutional subsampling (Conv-Sub) layer between the input embedding layer and the self-attention networks (SAN), a common technique in conventional E2E speech processing models based on continuous speech features.
This Conv-Sub layer reduces the sequence length to $\frac{1}{2}$ or even $\frac{1}{3}$ without performance degradation, as mentioned in~\cite{chang2023exploration}.
By default, a 2-layer 1D-convolutional (\textsc{CONV1D}) block is adopted.

\begin{table*}[htbp]
    \centering
    \caption{CER or WER (\%) on ASR benchmarks using the joint CTC/AED framework.}
    \vspace{-1em}
    \scalebox{0.8}{
    \begin{tabular}{l|cc|c|cc|cc||c}
        \toprule
        \multirow{2}{*}{Dataset} & \multirow{2}{*}{Language} & \multirow{2}{*}{Metric} & \multirow{2}{*}{Evaluation Sets} & \multicolumn{2}{c}{Discrete Tokenization} & \multicolumn{3}{c}{Results} \\
                                 &                           &                         &               &  K-means clusters & BPE size & FBank & Discrete Units & SSL (top line) \\
        \midrule
        AISHELL & CH & CER & dev/test & 2000 & 6000 &  4.2 / 4.5 & 4.6 / 4.9 & 3.8 / 4.0 \\
        CHiME4 & EN & WER & \{dt05,et05\}\_real\_5mics & 1000 & 2000 & 12.0 / 17.4 & \textbf{8.4 / 7.1} & 5.5 / 4.7 \\
        CommonVoice & FR & WER & dev / test & 1000 & 1500 & 12.8 / 14.6 & 20.9 / 23.5 & 12.9 / 15.0 \\
        Gigaspeech & EN & WER & dev / test & 1000 & 3000 & 11.8 / 11.6 & \textbf{11.2/11.6} & 10.2 / 10.3 \\
        How2-2000 & EN & WER & test & 2000 & 6000 & 10.9 & 10.9 & 10.8 \\
        LibriSpeech-100 & EN & WER & \{dev,test\}-\{clean,other\} & 2000 & 6000 & 6.1 / 16.7 / 6.3 / 17.0 & \textbf{3.8 / 6.7 / 3.9 / 7.0} & 3.2 / 5.3 / 3.1 / 5.5 \\
        LibriSpeech & EN & WER & \{dev,test\}-\{clean,other\} & 2000 & 6000 & 2.5 / 6.3 / 2.6 / 6.2 & \textbf{2.2 / 4.5 / 2.4 / 4.6} & 1.9 / 3.7 / 1.9 / 3.7 \\
        ML-SUPERB (1h) & 143 & CER & normal / few-shot & 2000 & 6000 & 59.6 / 58.6 & \textbf{44.8 / 46.4} & 21.8 / 38.9 \\
        Must-C & EN & WER & tst-COMMON & 2000 & 4000 & 8.3 & \textbf{6.1} & 5.7 \\
        SPGIspeech & EN & WER & \{dev,val\}\_unnorm & 1000 & 5000 & 6.0 / 6.0 & \textbf{5.9 / 5.9} & 5.5 / 5.5 \\
        SWBD & EN & WER & eval2000 (callhm / swbd) & 1000 & 1500 & 13.5 / 7.5 & \textbf{11.3 / 7.6} & 10.1 / 6.5 \\
        TEDLIUM3 & EN & WER & dev / test & 1000 & 2000 & 9.4 / 8.8  & \textbf{9.0 / 8.9} & 8.9 / 8.8 \\
        \bottomrule
    \end{tabular}
    }
    \vspace{-1em}
    \label{tab:asr_cer_wer}
\end{table*}
\begin{table*}[htbp]
    \centering
    \caption{Average sequence length (seq\_len) on training set of discrete speech unit (Disc\_Unit) and output asr transcription (ASR\_Trans). Lengths of discrete unit after manipulations are provided, including de-duplication $\rightarrow$ subword modeling $\rightarrow$ convolutional subsampling.}
    \vspace{-1em}
    \scalebox{0.83}{
    \begin{tabular}{l|c|ccc|cc|c}
    \toprule
    \multirow{2}{*}{Dataset} & \multirow{2}{*}{Disc\_unit} & \multicolumn{3}{c|}{Manipulations} & \multicolumn{2}{c|}{Conv-Sub} & \multirow{2}{*}{BPE-ASR\_Trans} \\
                             &                             & +De-dup & +BPE & length reduction ratio  & Type & length & \\
    \midrule
    AISHELL & 227 & 172 & 98 & 57\% & Conv1d2 & 49 & 14 \\
    CHiME4 & 383 & 260 & 178 & 54\% & \textbf{Conv1d1} & 178 & 106 \\
    CommonVoice (FR) & 236 & 177 & 153 & 65\% & Linear & 153 & 42 \\
    Gigaspeech & 217 & 150 & 90 & 59\% & Conv1d2 & 45 & 16\\
    How2-2000 & 297 & 207 & 127 & 57\% & Conv1d2 & 63 & 25\\
    LibriSpeech-100 & 639 & 455 & 257 & 60\% & Conv1d2 & 128 & 43 \\
    LibriSpeech & 619 & 435 & 247 & 60\% & Conv1d2 & 123 & 41 \\
    ML-SUPERB (1h) & 336 & 224 & 183 & 46\% & Conv1d2 & 92 & 66 \\
    MuST-C & 321 & 229 & 122 & 62\% & Linear & 122 & 25 \\
    SPGIspeech & 458 & 307 & 159 & 65\% & Conv1d2 & 79 & 30 \\
    SWBD & 269 & 182 & 144 & 46\% & Conv1d2 & 72 & 20 \\
    TEDLIUM3 &306 &208 &143 &53\% & Conv1d2 & 72 &36 \\
    \bottomrule
    \end{tabular}
    }
    \vspace{-1em}
    \label{tab:asr_length}
\end{table*}

\section{Experiments} \label{sec:exps}

\subsection{General Setup} \label{ssec:exp_setup}

Our experiments are conducted using the open-source E2E speech processing toolkit, ESPnet \cite{espnet}, aligning with~\cite{chang2023exploration}.
To thoroughly evaluate the performance of speech processing tasks using discrete speech representations, we adopted various tasks, including ASR, ST, and SLU.
For speech data processing and transcription normalization, we follow the existing recipes in ESPnet to make it comparable with prior works.

Unless stated otherwise, the joint CTC/AED framework is predominantly employed for both training and inference.
We use a 12-layer EBranchformer~\cite{kim2023branchformer} as the encoder architecture in all the experiments.
In each layer of the encoder, the hidden dimension $d$ is set to 256.
The number of heads in the self-attention is 4.
For the decoder, we employ a 6-layer Transformer decoder with 4 attention heads.
During inference, no external language models are used, and the beam size is set to 10 across all experiments, which may lead to slightly worse performance than other studies that primarily focus on achieving state-of-the-art performance.

The process of speech discretization hinges heavily on the choice of features.
In the prior study~\cite{chang2023exploration}, the final layer of the WavLM large model was adopted, guided by the learned weights from a pre-trained ASR model.
In this paper, we draw inspiration from a recent study~\cite{pasad2023comparative} using canonical correlation analysis (CCA), assessing the similarity between the layer representation and the word labels.
For example, both HuBERT and WavLM share similar patterns in their analyses due to their similar training objectives.
For monolingual experiments, we choose WavLM large for all English corpora \footnote{\url{https://huggingface.co/microsoft/wavlm-large}}, Chinese HuBERT large for Chinese corpus\footnote{\url{https://huggingface.co/TencentGameMate/chinese-hubert-large}},
and XLSR-53\footnote{\url{https://dl.fbaipublicfiles.com/fairseq/wav2vec/xlsr_53_56k.pt}} for French corpus.
Consequently, we choose layer 21 from large models, layer 9 from the base model and layer 11 from the XLSR-53 model, as they exhibit the highest CCA similarities with word labels.
For multilingual experiments on ML-SUPERB (143 languages), we use XLS-R-1b\footnote{\url{https://huggingface.co/facebook/wav2vec2-xls-r-1b}} and its layer 35 from empirical resynthesis analysis discussed in~\cite{barrault2023seamlessm4t}.
The number of K-Means clusters and the subword modeling vocabulary size are empirically selected considering the data variations and the length reduction balance.

\subsection{Automatic Speech Recognition}
\label{ssec:asr_results}

\subsubsection{Datasets} \label{sssec:asr_datasets}
To assess the efficacy of discrete speech units, we employed a diverse set of speech corpora, encompassing a wide range of acoustic characteristics: read English speech (LibriSpeech~\cite{panayotov2015librispeech}), noisy speech (CHiME4~\cite{vincent20164th}), telephony speech (SWBD~\cite{godfrey1992switchboard}), spontaneous speech (Gigaspeech~\cite{chen2021gigaspeech}, TEDLIUM3~\cite{hernandez2018ted}, How2~\cite{sanabria2018how2}, SPGIspeech~\cite{o2021spgispeech}, MuST-C~\cite{di2019must}), and non-English speech (AISHELL~\cite{bu2017aishell}, CommonVoice~\cite{ardila-etal-2020-common}, ML-SUPERB~\cite{shi23g_interspeech}).

\subsubsection{Results} \label{sssec:asr_results}
Table~\ref{tab:asr_cer_wer} provides a summary of the ASR results obtained using joint CTC/AED models.
We compare the peroformance of the ASR model using discrete units against conventionals ones using the FBank and SSL features.
Note that the same SSL representations are used to yield the discrete units.
We downsample both the FBank and SSL representations so that the frameshift is equivalent to 40ms.

Results reveal that the discrete units show commendable performance.
In most instances, the performance falls between FBank and SSL input, but it tends to align more closely with SSL input.
Comparing with the previous study~\cite{chang2023exploration}, using the 21st layer of the WavLM large model results in better performance on LibriSpeech, which further shows the importance of the discretization process.

\subsubsection{Efficiency of Discrete Units}
We summarize the sequence length information in the Table~\ref{tab:asr_length}. One benefit of using discrete units is that it can improve the training efficiency via the sequence length reduction. We can see that the length of the discrete unit after subword modeling is less than half of the SSL feature sequence. Thus, the training efficiency can be dramatically improved because of larger batch size and less IO overhead. In most of the discrete unit experiments, convolutional subsampling reduces input sequence length by half. However in CHiME4, halving the input sequence would violate the CTC assumption: the input sequence must not be shorter than the output sequence. For the CommonVoice and MuST-C, a linear layer leads to better convergence.

\begin{figure}[t]
    \centering
    \includegraphics[width=0.95\linewidth]{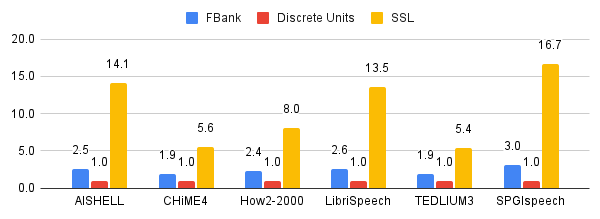}
    \caption{Illustration of training time per epoch using FBank / online SSL represetation / discrete units. We normalize the training time of discrete units to unit 1 for convenience.}
    \label{fig:training_time}
    \vspace{-1em}
\end{figure}

Fig.~\ref{fig:training_time} shows the training time for one epoch on several corpora, using FBank / online SSL feature / discrete units. It can be seen that the training time using discrete units is less than $50\%$ of that of using FBank. It's worth noting that online SSL feature extraction is computationally intensive but friendly to storage. 
Additionally, more aggressive subsampling (e.g. Conv1d3) can further improve the training speed without performance degradation in some cases\footnote{\url{https://github.com/espnet/espnet/tree/master/egs2/librispeech/asr2}}.

\begin{table}[htbp]
    \centering
    \caption{CER or WER (\%) of CTC and RNN-Transducer models.}
    \vspace{-1em}
    \scalebox{0.83}{
    \begin{tabular}{c|c|c||c}
    \toprule
    Dataset & FBank & Discrete Units & SSL (top line) \\
    \midrule
    \multicolumn{4}{c}{CTC} \\
    \midrule
    AISHELL & 5.8 / 6.2 & \textbf{5.3 / 5.5} & 3.9 / 4.2 \\
    LibriSpeech & 3.9 / 9.8 / 4.0 / 9.7 & \textbf{2.7 / 5.4 / 2.9 / 5.5} & 2.3 / 4.5 / 2.3 / 4.6  \\
    TEDLIUM3 & 10.2 / 10.3 & \textbf{8.8 / 9.6} & 8.4 / 10.0 \\
    \midrule
    \multicolumn{4}{c}{RNN-Transducer} \\
    \midrule
    AISHELL & 5.6 / 6.0 & 5.7 / 6.0 & 4.0 / 4.3 \\
    LibriSpeech & 2.5 / 6.2 / 2.7 / 6.2 & \textbf{2.2 / 4.6 / 2.4 / 4.5} & 2.0 / 4.1 / 2.1 / 4.3 \\
    TEDLIUM3 & 7.7 / 7.6 & \textbf{6.7 / 6.9} & 6.4 / 7.1 \\
    \bottomrule
    \end{tabular}
    }
    \label{tab:asr_ctc_transducer_cer_wer}
\vspace{-1em}
\end{table}

\subsubsection{ASR with Different Seq2Seq Models}

We further verified the efficacy on other alternative Seq2Seq models, namely, CTC and RNN-Transducer. Similar trends are found: using discrete units achieves the performance in between the FBank and SSL features, as shown in \Cref{tab:asr_ctc_transducer_cer_wer}.
\begin{table}[htbp]
    \centering
    \caption{LibriSpeech WER (\%) of different discrete units choices.}
    \vspace{-1em}
    \scalebox{0.81}{
    \begin{tabular}{c|c|c|c}
    \toprule
    Dataset & SSL & OWSM & EnCodec \\
    \midrule
    LibriSpeech & \textbf{2.2 / 4.5 / 2.4 / 4.6} & 3.9 / 10.1 / 4.2 / 10.2 & 3.2 / 8.5 / 3.0 / 8.5 \\
    \bottomrule
    \end{tabular}
    }
    \vspace{-1em}
    \label{tab:feats}
\end{table}

\subsubsection{ASR with Different Types of Discrete Units} \label{sssec:asr_different_units}

So far, SSL representations are used to derive discrete units, as they were known to be very robust in different tasks \cite{baevski2020wav2vec,chen2022wavlm}. However, other large models and discrete unit approaches are recently emerging, such as Whisper~\cite{radford2023robust} and neural codec models~\cite{zeghidour2021soundstream,defossez2022high}. We adopt a Whisper-like model trained by ESPnet \footnote{\url{https://github.com/espnet/espnet/pull/5120}} and the publicly available EnCodec~\cite{defossez2022high} model. For EnCodec, we use all 8-level discrete tokens at each frame, summing up embeddings of 8-levels per frame.
The embedding layer is initialized from the pre-trained codebook and frozen, followed by a layer normalization for stability.
Subword modeling cannot be applied in this case.
Table~\ref{tab:feats} shows our preliminary results.
However, we observe that the SSL-based method still has the upper hand over the others.
Nevertheless, we aim to continue exploring various settings in our future work.

\subsection{Speech Translation Results} \label{ssec:st_results}
\begin{table}[]
    \centering
    \caption{BLEU scores of speech translation tasks.}
    \vspace{-1em}
    \scalebox{1.0}{
    \begin{tabular}{c|c|c||c}
    \toprule
    Dataset & FBank & Discrete tokens & SSL (top line) \\
    \midrule
    MuST-C En-De & 26.7 & \textbf{28.6} & 29.7 \\
    MuST-C En-Es & 31.2 & \textbf{33.0} & 33.7 \\
    MuST-C En-Fr & 37.1 & \textbf{38.7} & 40.2 \\
    \bottomrule
    \end{tabular}
    }
    \label{tab:st_bleu}
\end{table}

Table~\ref{tab:st_bleu} shows the ST results; these models use CTC / attention following~\cite{yan2023ctc, yan-etal-2023-espnet}.
Similar to the ASR experiments in \Cref{ssec:asr_results}, the performance of ST using discrete units is slightly worse than that using the SSL features but better than FBank.

\subsection{Spoken Language Understanding Results} \label{ssec:slu_results}

\begin{table}[]
    \centering
    \caption{Intent classification accuracy (\%) of SLURP dataset.}
    \vspace{-1em}
    \scalebox{1.0}{
    \begin{tabular}{c|c|c||c}
    \toprule
    Dataset & FBank & Discrete tokens & SSL \\
    \midrule
    SLURP & 86.9 / 86.3 & 81.8 / 80.8 & 84.2 / 83.3 \\
    \bottomrule
    \end{tabular}
    }
    \vspace{-1em}
    \label{tab:slu_acc}
\end{table}

We conducted experiments on an SLU task using the SLURP~\cite{bastianelli2020slurp} dataset. The intent classification accuracy is presented in Table~\ref{tab:slu_acc}. Unlike the other experiments, in this preliminary result, we did not observe an improvement in both SSL continuous features and discrete tokens. 
We hypothesize that the 21st layer of the WavLM large selected following the ASR experiments may not be optimal for SLU purposes due to the dependency of the optimal layer on the downstream task, as reported in~\cite{pasad2023comparative,baevski2020wav2vec}.

\section{Conclusion} \label{sec:concl}

This paper explores the efficacy of incorporating discrete speech units as inputs across a spectrum of speech processing tasks, encompassing ASR, ST, and SLU. To evaluate the versatility of discrete units in diverse scenarios, we conducted experiments on datasets with varying characteristics. Drawing inspiration from canonical correlation analysis (CCA), we improved our choice of self-supervised learning (SSL) features, resulting in a noticeable performance enhancement. Consequently, the utilization of discrete units not only outperforms FBank features but also substantially enhances efficiency. These findings underscore the promise of discrete unit input in speech processing. Future research avenues could delve into investigating alternative discretization techniques.


\section{Acknowledgements}
Some experiments of this work used the Bridges2 system at PSC and Delta system at NCSA through allocation CIS210014 from the Advanced Cyberinfrastructure Coordination Ecosystem: Services \& Support (ACCESS) program, which is supported by National Science Foundation grants \#2138259, \#2138286, \#2138307, \#2137603, and \#2138296. We also gratefully acknowledge the support of NVIDIA Corporation with the donation of the A6000 GPUs used for this research.

\clearpage

\section{References}
{
\printbibliography
}
\end{document}